\documentclass[lettersize,journal]{IEEEtran}
\usepackage{amsmath,amsfonts}
\usepackage{array}
\usepackage[caption=false,font=normalsize,labelfont=sf,textfont=sf]{subfig}
\usepackage{subfig}
\usepackage{textcomp}
\usepackage{stfloats}
\usepackage{url}
\usepackage{verbatim}
\usepackage{graphicx}
\usepackage{cite}
\usepackage{bbding}
\usepackage{booktabs}
\usepackage{color}
\usepackage{xcolor}
\usepackage{colortbl}
\usepackage{amsfonts}
\usepackage{amsmath}
\usepackage[ruled,linesnumbered]{algorithm2e}
\usepackage{newfloat}
\usepackage{listings}
\usepackage[colorlinks,urlcolor=blue,linkcolor=blue,citecolor=blue]{hyperref}
\usepackage{orcidlink}

\begin{document}

\title{In-Loop Model Adaptation with Coupled Latent-Noise Guidance for High-Fidelity Subject-Driven Text-to-Image Generation}

\author{Yushun Tang$^{\orcidlink{0000-0002-8350-7637}}$, Weiming Chen$^{\orcidlink{0000-0002-0586-1278}}$, Siyi Liu$^{\orcidlink{0009-0001-4263-535X}}$, Yi Zhang$^{\orcidlink{0000-0002-5831-0170}}$, \\ Feng Wu$^{\orcidlink{0000-0001-7266-5579}}$,~\IEEEmembership{Fellow, IEEE}, and Zhihai He$^{\orcidlink{0000-0002-2647-8286}}$,~\IEEEmembership{Fellow, IEEE}
\thanks{This work was supported by the National Natural Science Foundation of China (No. 62331014), Project 2021JC02X103, and the Center for Computational Science and Engineering at Southern University of Science and Technology. \textit{(Corresponding author: Zhihai He.)}}

\thanks{
Yushun Tang, Weiming Chen, Siyi Liu, and Yi Zhang are with the Department of Electrical and Electronic Engineering, Southern University of Science and Technology, Shenzhen 518055, China.}

\thanks{
Feng Wu is with the School of Information Science and Technology, University of Science and Technology of China, Hefei 230027, China.}

\thanks{Zhihai He is with the Department of Electrical and Electronic Engineering, Southern University of Science and Technology, Shenzhen 518055, China, and also with Pengcheng Lab, Shenzhen 518066, China (e-mail: hezh@sustech.edu.cn).}



}

\markboth{IEEE Transactions on Multimedia}%
{Shell \MakeLowercase{\textit{et al.}}: A Sample Article Using IEEEtran.cls for IEEE Journals}


\maketitle

\begin{abstract}
Text-to-image diffusion models have achieved remarkable success in generating high-quality images from a given text prompt. Subject-driven generation aims to synthesize customized images to mimic the appearance of subjects in given reference images within different visual contexts specified by the text prompts. The central challenge here is that, when the reference image changes, the diffusion model cannot efficiently adapt to different visual contexts while consistently maintaining the subject identity. Existing methods either train the model with a large domain-specific dataset or fine-tune the model using the reference image for hundreds of iterations before actual image generation. In this work, we explore a new approach, called \textit{In-Loop Model Adaptation} (IMA), which adapts the core diffusion model at each generation step during the actual process of image generation, without being trained on the reference image before the generation process. To this end, we establish a DDIM inversion chain that maps the reference image to a sequence of latent, as well as a text-to-image generation chain which generates the image from the text prompt only. We then introduce a masked latent consistency loss and a noise regularization loss to characterize the latent-noise difference between the diffusion model and these two chains at each generation step. This coupled latent-noise loss is used to guide the in-loop model adaptation to preserve the subject identity specified by the reference image while maintaining accurate alignment with the text prompt, resulting in high-fidelity text-to-image generation. Our extensive experiments demonstrate that our proposed IMA method significantly improves the performance of subject-driven text-to-image generation.
\end{abstract}


\section{Introduction}
\label{sec:intro}
Recently developed large text-to-image models, such as Stable Diffusion \cite{rombach2022high}, have demonstrated powerful capabilities to generate images from text prompts \cite{jiang2024anime,qing2024diff}. Subject-driven generation \cite{ruiz2023dreambooth,textual_inversion,custom-diffusion,Ip-adapter,Elite,li2024blip,zhang2024ssr,xu2024sgdm} aims to synthesize customized images in various contexts based on a few reference images while preserving the subject's identity. Despite the remarkable success in text-to-image diffusion models, existing methods are still facing the significant challenge of consistently preserving the subject identity specified by the reference image while achieving accurate alignment with the text prompt.

\begin{figure}[!t]
    \centering
    \includegraphics[width=\linewidth]{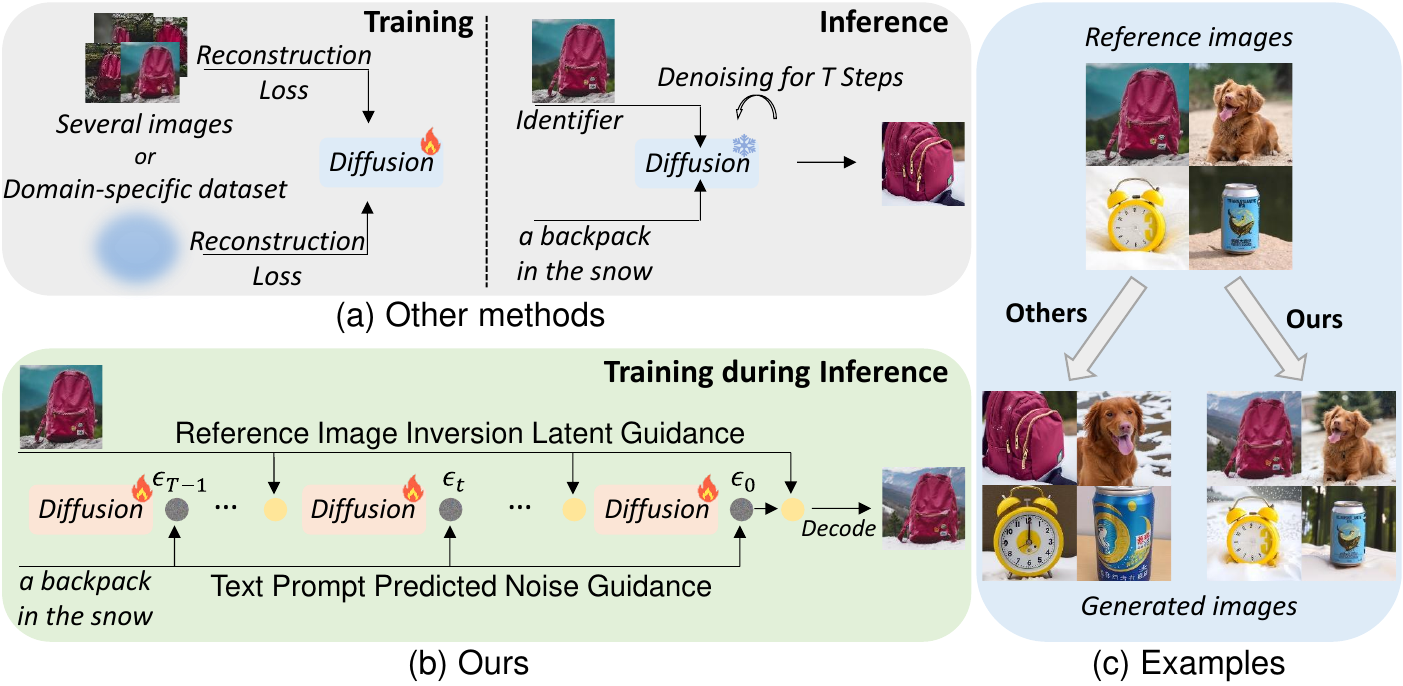}
    \caption{{Comparison between other methods and our method. (a) Existing methods finetune the diffusion U-Net on several reference images or a large domain-specific dataset during the training stage. Then, given the text prompt and a reference image or identifier, they generate new images with the finetuned model during inference. (b) In our approach, the weight parameters of the Key and Value in the image cross-attention module layers are fine-tuned online during the inference stage, guided by the coupled latent-noise chains. (c) The generated examples illustrate that other methods often fail to preserve subject identity consistency across different visual contexts, thereby visually highlighting the central challenge addressed in this paper.}}
    \label{fig:intro}
\end{figure}

During our studies, we observed that the diffusion model pre-trained on large-scale datasets suffers from the so-called generalization problem in subject-driven generation: Given the reference image, the diffusion model cannot efficiently adapt to different visual contexts while consistently maintaining the subject identity. 
In other words, there is a significant domain gap between the pre-trained diffusion model and the real-world reference image in subject-driven generation. 
To overcome this domain gap, similar to domain adaptation and domain generalization in transfer learning, two major approaches have been explored: (1) finetuning-based methods  \cite{ruiz2023dreambooth,textual_inversion,custom-diffusion} which train model in a few reference images of the target domain, and (2) finetuning-free methods \cite{Ip-adapter,Elite,li2024blip,zhang2024ssr} which train model in a large domain-specific dataset of the source domain. The finetuning-based methods require hundreds of steps to train the model for each new subject, making it very difficult to scale up for real applications.  For example, the Textual Inversion approach \cite{textual_inversion} fine-tunes the additional text embedding vector, and the DreamBooth approach \cite{ruiz2023dreambooth} fine-tunes the entire diffusion U-Net model with a unique identifier for hundreds of iterations given a few reference images. 
In contrast, finetuning-free methods train the diffusion model on domain-specific images and generate new images conditioned by reference images and text prompts in a zero-shot manner. For instance, the IP-Adapter \cite{Ip-adapter} trains an image encoder in a dataset including about $10$ million text-image pairs. The ELITE method \cite{Elite} expands to hierarchical global and local mapping in a dataset containing $125$k images with $600$ object classes. However, these methods work well for a narrow range of subjects and struggle to generalize to arbitrary subjects. Both approaches require computationally intensive training before actual image generation, either during training in the source domain with a large dataset or re-training in the target domain with reference images before the actual image generation process. A research question arises: \textit{Can we perform online adaptation of the diffusion model during the actual process of text-to-image generation to produce high-fidelity images?}

With this motivation, in this work, we propose a new approach called \textit{In-Loop Model Adaptation} (IMA), which adapts the core diffusion U-Net model during the text-to-image generation process. This adaptation process involves fine-tuning the Key-Value weights of the cross-attention modules, which are crucial for the image generation process. As shown in Figure \ref{fig:intro}, we adapt the diffusion U-Net during inference guided by the coupled latent-noise consistency.
{We establish a DDIM inversion chain that maps the reference image to a sequence of latent, as well as a text-to-image generation chain, which generates the image from the text prompt only.}
The in-loop online adaption is guided by the feedback from a latent-noise consistency verification. Specifically, we enhance subject identity preservation through masked latent consistency between the DDIM inversion \cite{songdenoising} chain path and the generating path. This consistency ensures that the generated images retain the distinctive features of the subjects in the reference image. In the meanwhile, we improve text prompt alignment by introducing a noise regularization between the generating path and the text-to-image chain path without reference images. This regularization loss ensures that the generated images accurately align with the textual prompts.

The \textbf{major contributions} of this work can be summarized as follows: (1) 
We first propose an innovative inference-time In-Loop Model Adaptation method that adapts the Key-Value weight parameters in the image cross-attention modules of the diffusion U-Net model, guided by coupled latent-noise path consistency during subject-driven generation.
(2) We introduce masked latent consistency between the DDIM inversion chain path and the generating path, significantly improving the preservation of subject identity.
(3) We enhance text prompt alignment by incorporating predicted noise regularization between the generating path and the text-to-image chain path without reference images.
(4) We demonstrate the effectiveness of our approach through comprehensive experimental results, showing significant improvements in subject-driven text-to-image generation.

\section{Related Work}
This work is related to text-to-image diffusion models, subject-driven generation, and online test-time adaptation.

\paragraph{Text-to-Image Diffusion Models}
Image generation methods that generate images conditioned on text description are mainly based on GANs \cite{GAN} and Diffusion models \cite{rombach2022high,nichol2022glide,rombach2022high}. Recently, Diffusion models have achieved state-of-the-art synthesis results in terms of image quality and diversity than previous GAN-based and autoregressive image generation models.
Recent works such as GLIDE \cite{nichol2022glide}, DALL-E2 \cite{dalle2}, Imagen \cite{imagen}, and Stable Diffusion \cite{rombach2022high} have achieved generating diverse and high-quality images that match the arbitrarily complex text prompt. Among them, DALL-E2 employs a diffusion model conditioned image embedding, and a prior model was trained to generate image embedding by giving a text prompt. DALL-E2 not only supports text prompt for image generation but also image prompt. Both GLIDE \cite{nichol2022glide} and Imagen \cite{imagen} use cascaded diffusion models to first generate low-resolution images and then produce high-resolution images. Stable Diffusion is based on latent diffusion models (LDM) \cite{rombach2022high} and opts to conduct conditional text-to-image diffusion in a latent space of reduced dimensionality to achieve faster training and sampling. None of these models offer fine-grained control or effectively preserve the details of the image subject during new image generation.

\paragraph{Subject-Driven Generation}
Subject-driven text-to-image generation aims to generate the given subject in a novel context based on text prompts. Some methods use finetuning-based approaches to embed subjects into diffusion models. Textual Inversion \cite{textual_inversion} optimizes the embedding to map a specific subject to a placeholder text embedding.
DreamBooth \cite{ruiz2023dreambooth} employs a similar methodology but additionally fine-tunes the diffusion model, enabling it to more accurately capture and reproduce the unique visual characteristics of the subject. Custom Diffusion \cite{custom-diffusion}, on the other hand, optimizes only the parameters within the cross-attention layers of the text-to-image diffusion model, effectively learning new concepts. When dealing with the composition of multiple concepts, this method can independently fine-tune each concept model and then integrate them into a unified model through constrained optimization techniques. One known drawback of these methods is extensive fine-tuning and optimization for each subject. To address this concern, tuning-encoder \cite{tuning-encoder} reduces the fine-tuning time overhead by generating a set of latent codes through diffusion inversion. 
A number of finetuning-free works that do not need training in the given reference image, have explored using additional networks to inject reference image information for subject generation. IP-Adapter \cite{Ip-adapter} introduces a decoupled cross-attention mechanism to achieve a more effective image prompt adapter that can be generalized to other custom models fine-tuned from the same base diffusion model. ELITE \cite{Elite} and InstantBooth \cite{shi2024instantbooth} use global and local mapping networks to project reference images into word embeddings and inject reference image patch features into cross-attention layers to enhance local details. BLIP-Diffusion \cite{li2024blip} introduced a new pre-trained encoder that supports multimodal control.
However, these methods are limited to a narrow range of subjects and struggle to generalize to generic subjects well. Both the finetuning-based and the finetuning-free approaches are computationally intensive, either during pre-training on the source domain or when fine-tuning with reference images in the target domain. 
In comparison to existing methods which need to train the model using reference images before the actual generation process, our IMA method is able to adapt the diffusion model within the process of image generation to ensure high-fidelity text-to-image generation while maintaining low computational complexity.

\begin{figure*}[!htbp]
    \centering
    \includegraphics[width=0.92\linewidth]{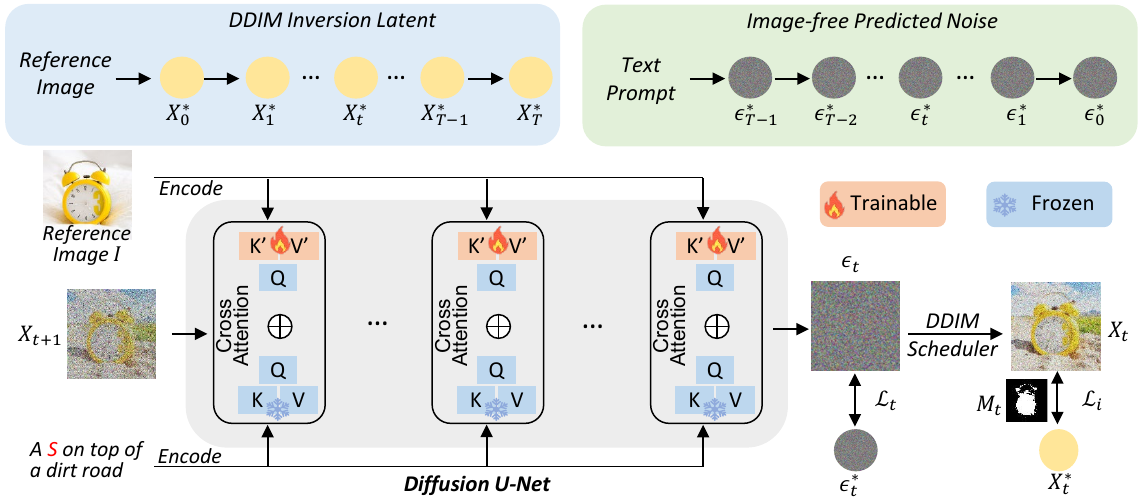}
    \caption{An overview of the proposed In-Loop Model Adaptation with Coupled Latent-Noise Guidance for subject-driven text-to-image generation. The DDIM Inversion latent guidance and Image-free predicted noise guidance (\textbf{Top}). The Key-Value weight parameters of the image cross-attention module in diffusion U-Net are updated online and guided by the above two loss functions during the generation process (\textbf{Bottom}). The weight parameters of the next step are from the updated parameters in the current step.}
    \label{fig:overview}
\end{figure*}

\paragraph{Online Test-time Adaptation}
This work is also related to online Test-time Adaptation. Online Test-time Adaptation (TTA) is designed to align a pre-trained model with new, unlabeled data that exhibits domain shifts during the inference phase \cite{liang2024comprehensive,tang2024learning,tang2024domain,tang2025dual,wen2023test}. Various methods have been developed to achieve this alignment. For instance, the TENT method \cite{wang2020tent} proposes fully test-time adaptation by recalibrating Batch Normalization layers in real-time. The NHL method \cite{tang2023neuro} enhances early-layer feature representations before prediction through unsupervised learning. The method proposed by \cite{gan2023decorate}  adjusts target inputs by learning image-specific visual prompts during the testing stage.
All of these TTA methods fine-tune the pre-trained model online during inference to enhance the model's generalization ability in the target domain. Similarly, some image generation methods also involve test-time fine-tuning in the target domain. {For example, A-STAR \cite{agarwal2023star} trains the latent code during generation by attention segregation and attention retention, for a semantically closer to the text prompt.} The NTI method \cite{NTI} trains the null text embedding for multiple iterations to invert the image with a meaningful text prompt into the pre-trained model’s domain, for high-fidelity editing of real images.
In this work, we observe that the pre-trained diffusion U-Net similarly struggles to generalize well to new reference images in subject-driven text-to-image generation. Therefore, we investigate inference-time adaptation for the diffusion U-Net to address this challenge in order to achieve high-fidelity image generation.


\section{Method}
In this section, we present our method of In-Loop Model Adaptation for subject-driven text-to-image generation.

\subsection{Method Overview}
Subject-driven generation aims to synthesize customized images $I_t$ with user prompt $P$ based on a given reference image $I$. Formally, $I_t = \mathcal{M}_\theta(I, P)$, where $\mathcal{M}_\theta$ represents the Stable Diffusion models including text encoder, VAE, diffusion U-Net, and scheduler. In this work, we propose an online in-loop adaptation for the diffusion U-Net during the image generation process. The current Key-Value weight parameters of the image cross-attention modules $\theta_t^{kv}$ are updated based on the last predicted noise $\epsilon_{t+1}$ and latent $X_{t+1}$ by the scheduler, guided by the coupled latent-noise consistency.

Specifically, for a given reference image $I$ and a user query text prompt $P$, we first invert the reference image $I$ to obtain the  latent chain $\mathbf{X^*} = \{X_0^*, X_1^*, \cdots, X_t^*, \cdots, X_{T-1}^*, X_T^*\}$ by DDIM inversion \cite{songdenoising}. This image inversion chain is used as the image model guidance for the subsequent generation process, as the masked inverted latents represent the target subject. During the inference stage, we incorporate another diffusion denoising chain $\mathcal{E}^* = \{\epsilon_{T-1}^*, \cdots, \epsilon_t^*, \cdots, \epsilon_1^*, \epsilon_0^*\}$ without reference image. This single text-to-image chain without reference image is utilized as another model guidance, as it can generate images that align well with the given text prompt. The Key-Value weight parameters of the image cross-attention module $\theta_t^{kv}$, i.e., $W_K'$ and $W_V'$, are adapted during inference for $T$ steps. At time-step $t$ of the generation process, the predicted noise $\epsilon_t$ is guided by $\epsilon_t^*$ and the latent $X_t$ is guided by $M \odot X_t^*$, where $M$ represents the subject mask. 
As shown in Figure \ref{fig:overview}, the proposed in-loop adaptation enhances both subject identity and text prompt alignment guided by the above coupled latent-noise chains. In the following sections, we will introduce our proposed method in more detail.

\begin{figure*}[!b]
    \centering
    \includegraphics[width=\linewidth]{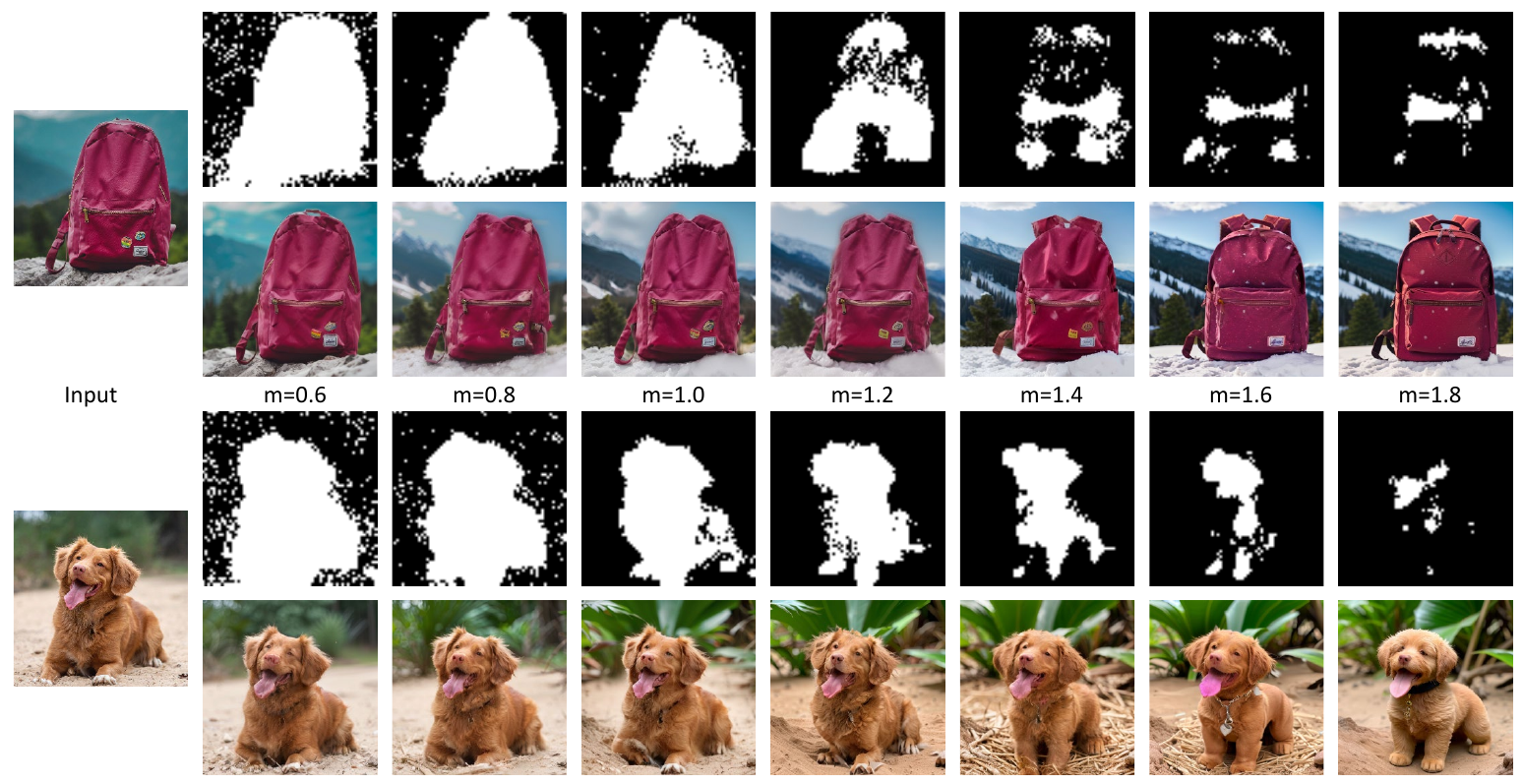}
    \caption{A subject-driven generation example for different $m$ value in equation (\ref{eq:m_mask}). The images in the leftmost are the input reference images. All other images are generated in the text prompt ``\textit{A backpack in the snow}" (Top) and ``\textit{A dog in the jungle}" (Bottom). {The corresponding masks $M_T$ in the last step are also visualized.} As the value of $m$ increases, the mask area is reduced and the consistency of details between the reference image and the generated image tends to reduce.}
    \label{fig:m_mask}
\end{figure*}

\subsection{Diffusion-Based Subject-Driven Generation}
Diffusion models consist of two processes: a diffusion process, which gradually adds Gaussian noise to the image over $T$ steps, and a denoising process which gradually removes Gaussian noise from images with a learnable U-Net model.
Building upon the success of the diffusion models, text-to-image diffusion models have proven capable of generating high-quality images from text prompts. The text-conditioned diffusion U-Net model $\epsilon_{\theta}$ is trained on the latent space to predict the noise in each time-step based on the text prompt condition. The mean-squared loss function for training the latent diffusion model is:
\begin{equation}
    \mathcal{L}_{LDM} := \mathbb{E}_{X_0, \epsilon \sim \mathcal{N}(\textbf{0},\textbf{I})}\parallel \epsilon - \epsilon_\theta(X_t, t, c) \parallel_2^2,
\end{equation}
where time-step $t \in [0, T]$ and $c$ represents the text prompt condition encoded by a pre-trained text encoder. Once the diffusion U-Net is trained, a random Gaussian noise $X_T$ is able to denoise to image latent in an iterative manner during the inference stage. For the text conditional diffusion models, cross-attention is adopted in Stable Diffusion U-Net to incorporate text information in the denoise process. Specifically, given the latent image feature $f$ with dimension $d$, the cross-attention is computed as follows:
\begin{equation}
    \text{Attention}(Q,K,V) = \text{Softmax}(\frac{QK^{\top}}{\sqrt{d}})V,
\end{equation}
where $Q = f\cdot W_q, K = c \cdot W_k, V = c \cdot W_v$. $W_q, W_k$, and $W_v$ are weight parameters of query, key, and value projection layers, respectively.

{For subject-driven generation, both the text prompt feature $c_t$ and the reference image feature $c_i$ are needed for the generation process.} we follow existing methods \cite{Ip-adapter,Elite} to insert another image cross-attention layer conditioned by the image feature to the original cross-attention layers. Specifically, for image Key $K'$ and image Value $V'$, the new output for the combined cross-attention is given by:
\begin{equation}
\begin{aligned}
    &\text{Attention}(Q,K,V,K',V') \\
    = &\text{Softmax}(\frac{QK^{\top}}{\sqrt{d}})V + \lambda \cdot \text{Softmax}(\frac{QK'^{\top}}{\sqrt{d}})V',
\end{aligned}
\end{equation}
where $K'= c_i \cdot W'_k, {V' = c_i \cdot W'_v}$. $W'_k$ and $W'_v$ are weight parameters of key and value projection layers, respectively.
It has been recognized that cross-attention layers play a crucial part in the tuning effort to improve generalization for domain tuning \cite{gal2023encoder}.
{In this work, we perform test-time adaptation on the Key-Value projection parameters, denoted as $W'_k/W'_v$, of the additional image cross-attention modules attached to the U-Net backbone.} This {adaptation} process is guided by the coupled latent-noise chains, as illustrated in section \ref{sec:ddim} and \ref{sec:t2i}, to ensure the adaptation is well aligned with the reference image and text prompt to achieve high-fidelity image generation.

\subsection{DDIM Inversion Latent Guidance}
\label{sec:ddim}
Denoising Diffusion Implicit Models (DDIM) Inversion is a technique used in generative modeling to reverse the diffusion process and obtain an inverted latent representation chain from a given image \cite{dhariwal2021diffusion,songdenoising}. The inverted latent representation chain includes the information and identity of the original reference image.
To ensure that the identity of the subject from the reference image is preserved better, we introduce a masked latent consistency mechanism between the DDIM inversion chain path and the generating path.  By applying a binary mask to the latent space, we can selectively focus on critical features that define the subject’s identity, allowing for more accurate and consistent subject representation in the generated images. In this work, the mask $M$ is automatically generated by selecting the high correlation area between the category word and the image patch features during the generating process. Specifically, at each time step $t$, the latent mask $M_t$ is defined as:
\begin{equation}\label{eq:m_mask}
    M_t = [A_t > m \times \text{Mean}(A_t)],
\end{equation}
where $A_t$ is the attention map and $m$ is a hyper-parameter to control how much subject area should be preserved. When $m$ is small, the attended area will be large, and the generated image will retain more information from the reference image. Otherwise, when $m$ increases, the generated image will preserve less of the subject identity. An example is shown in Figure \ref{fig:m_mask}. We can see that when $m=0.6$, the generated image closely resembles the input reference image. However, when $m=1.8$, the generated image shows less preservation of the input reference image.
Therefore, the masked DDIM inversion latent consistency can be denoted as:
\begin{equation}
    \mathcal{L}_i = \parallel M_t \odot X_{t}^* - M_t\odot X_{t} \parallel_2^2,
\end{equation}
where $X_{t}^*$ represents the latent representation obtained from the DDIM inversion chain path, and $X_{t}$ represents the latent representation from the generating path. The binary mask $M_t$ is applied element-wise to both latent representations to focus on the critical features. By minimizing this loss, we ensure that the generated image retains the distinctive features of the reference image, thus preserving the subject's identity.

\begin{figure}[!ht]
    \centering
    \includegraphics[width=\linewidth]{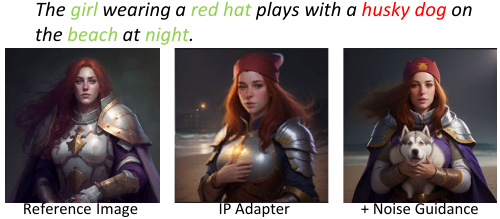}
    \caption{Given a reference image and text prompt, the IP-Adapter method ignores the text constraint ``\textit{husky dog}". Once incorporating the proposed predicted noise guidance, the method successfully align to all the text prompts.}
    \label{fig:noise_guidance}
\end{figure}

\subsection{Image-Free Predicted Noise Guidance}
\label{sec:t2i}
\begin{figure*}[!ht]
    \centering
    \includegraphics[width=\linewidth]{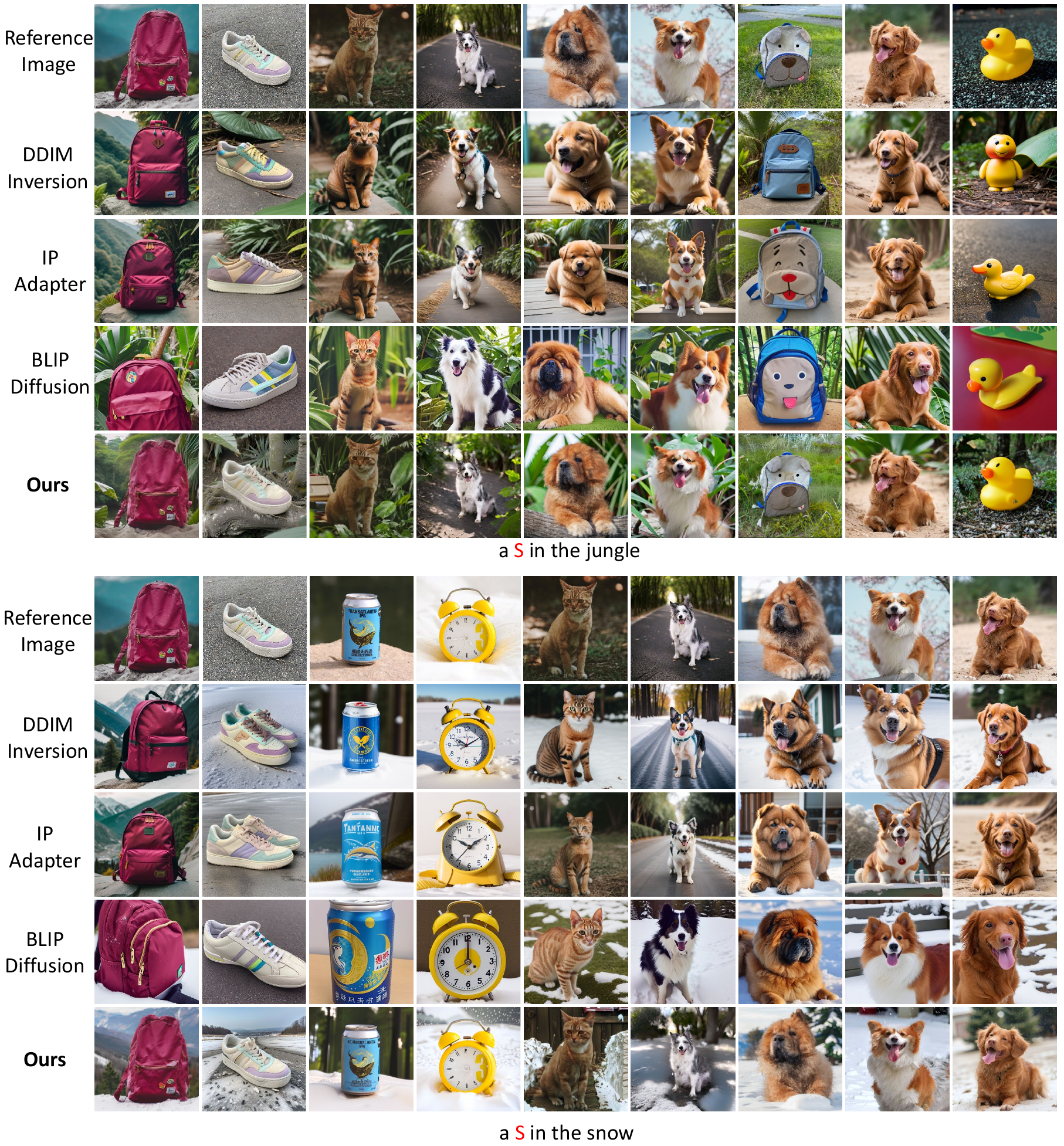}
    \caption{Representative examples for the subject-driven generation of the generated images in the DreamBench dataset. Our method demonstrates better subject identity and text alignment compared to DDIM Inversion, IP-Adapter, and BLIP-Diffusion.}
    \label{fig:jungle_snow}
\end{figure*}

In our experiment, we find that when inserting new image cross-attention layers into the original diffusion U-Net, the consistency between the generated images and text prompt is degraded, especially in text prompts including multiple constraints, as shown in Figure \ref{fig:noise_guidance}. This is because adding the image cross-attention features to the text cross-attention could change the original text's semantic feature. In Figure \ref{fig:jungle_snow}, the images generated by IP-Adapter and BLIP-Diffusion both ignore some text words in some cases, such as the last column, the IP-Adapter method and the BLIP-Diffusion method both ignored the word ``\textit{jungle}". In contrast, the DDIM Inversion, which generates images using the inverted initial noise latent without adding image cross-attention, aligns well with the text prompt. To improve text prompt alignment, we implement predicted noise regularization between the generating path and the original vanilla text-to-image chain path without a reference image. This regularization process involves comparing the noise predicted during the generation with the noise from the original text-to-image chain path, enabling better adherence to the text prompt while generating new images. The noise regularization loss is defined as follows:
\begin{equation}\label{loss:ipng}
    \mathcal{L}_t = \parallel \epsilon_{t}^* - \epsilon_{t} \parallel_2^2,
\end{equation}
where $\epsilon_{t}^*$ represents the noise predicted by the original text-to-image chain path, and $\epsilon_{t}$ represents the predicted noise in the subject-driven generation path. In the $t$-th step, given the pre-trained diffusion U-Net model $\epsilon_\theta$, the image-free predicted noise $\epsilon_{t}^*$ can be formulated as:
\begin{equation}
    \epsilon_{t}^* =  \epsilon_{\theta}(X_{t+1}, t, c_p),
\end{equation}
where $c_p$ is the embedding of the text prompt $P$.
By minimizing equation (\ref{loss:ipng}), we make sure that the noise patterns in the generating path align with those in the original chain path, resulting in better adherence to the text prompt. The noise regularization loss is incorporated into the overall loss function used during the generation process. This ensures that the model simultaneously optimizes for both subject identity preservation (through masked latent consistency) and text prompt alignment (through noise regularization).

\begin{algorithm}[!htbp]
\caption{Pseudo code of the algorithm.}
\label{alg: algorithm}
\KwIn {Pre-trained diffusion U-Net model $\epsilon_{\theta}$; target reference image $I$; text prompt $P$.}
\KwOut {Synthesized Image $I_t$.}
Initialize the diffusion U-Net model $\epsilon_{\theta}$ with the pre-trained parameter weights; learning rate $\eta > 0$; \\
$\mathbf{X^*} \leftarrow$ DDIM inversion over reference image $I$\\
Initilalize $X_T = X_T^*$, embedding $c_p$ for $P$\;
\For {time-step $t = T-1, \cdots, 0$}{
$\epsilon_{t}^* =  \epsilon_{\theta}(X_{t+1}, t, c_p)$\; 
$\epsilon_{t} =  \epsilon_{\theta}(X_{t+1}, t, c_p + \lambda c_i)$\;
$X_{t} = Scheduler(X_{t+1}, \epsilon_t)$\;
$\mathcal{L} = \alpha||\epsilon_{t}^* - \epsilon_{t}||_2^2 + \beta||M_t\odot X_{t}^* - M_t\odot X_{t}||_2^2$\;
Update $\tilde{\theta}_t^{kv} \leftarrow \tilde{\theta}_{t+1}^{kv}-\eta \nabla_\theta \mathcal{L}$\;
 }
 Output $I_t$ decoded by $X_0$.
\end{algorithm}

\begin{figure*}
    \centering
    \includegraphics[width=\linewidth]{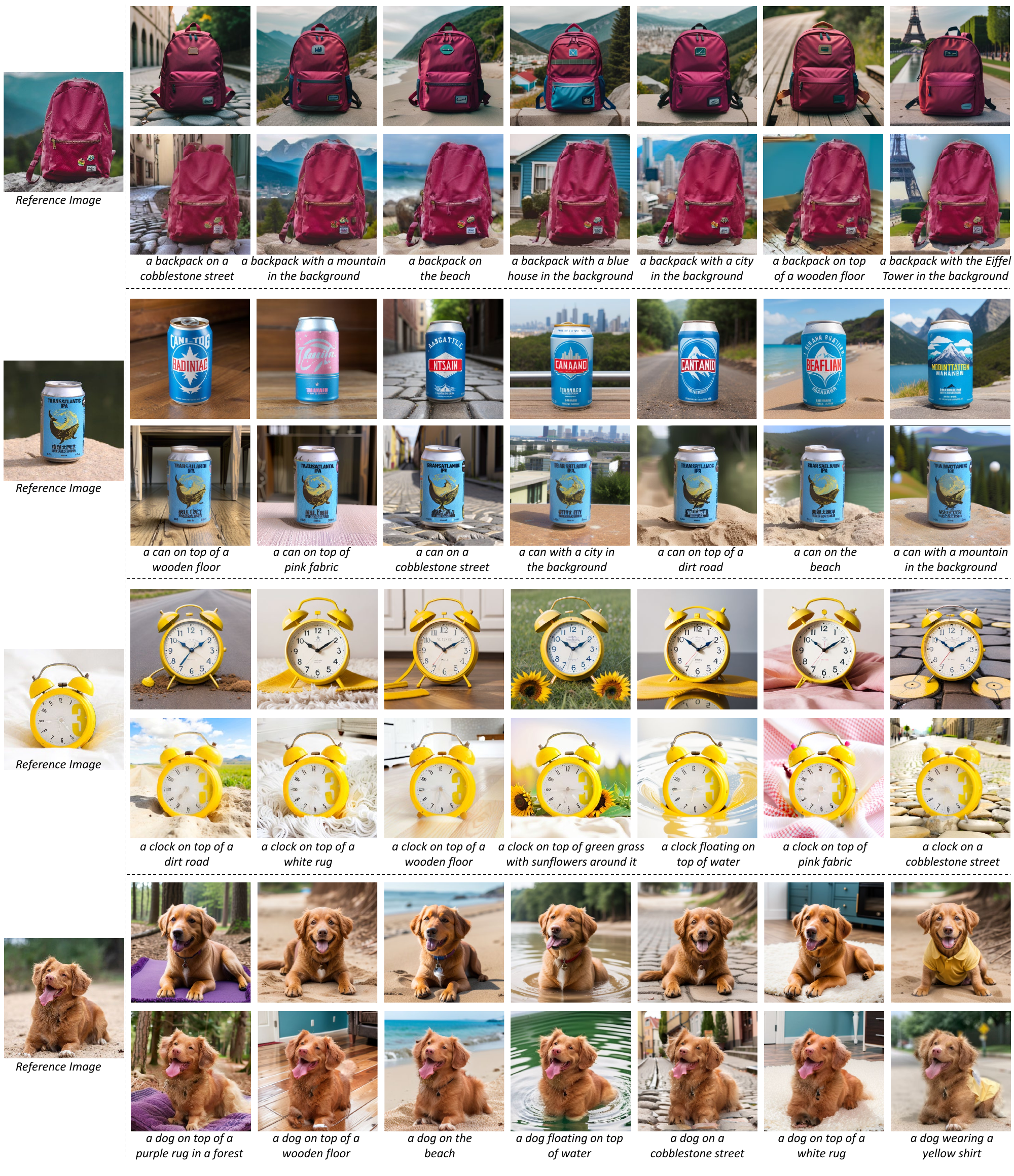}
    \caption{Additional examples for the subject-driven generation across various text prompts. The leftmost images for each row are the input reference images and the others are subject-driven generated images using IP-Adapter (\textbf{Top}) and Ours (\textbf{Bottom}).}
    \label{fig:supp_prompts}
\end{figure*}

Overall, the Key-Value weight parameters of image cross-attention in the Diffusion U-Net model are adapted during inference guided by the above two loss functions. The total loss function can be denoted as:
\begin{equation}\label{eq:loss}
    \mathcal{L} = \alpha \mathcal{L}_t + \beta \mathcal{L}_i,
\end{equation}
where $\alpha$ and $\beta$ are balance hyper-parameters whose values are set to $1.0$ in default.
The pseudo-code of our proposed method is provided in Algorithm \ref{alg: algorithm}.

\section{Experimental Results}
\label{sec:exp}
In this section, we conduct experiments on dataset benchmark to evaluate the performance of our proposed In-Loop Model Adaptation for subject-driven generation.

\subsection{Benchmark Dataset and Baselines}
In our experiments, we use the widely used Dreambench \cite{ruiz2023dreambooth}, a dataset consisting of 30 subjects, including unique objects and pets such as backpacks, stuffed animals, dogs, cats, sunglasses, cartoons, etc. We randomly choose one image for each subject as the reference image. We also use the same 25 prompts for fair comparison. We compare our proposed IMA method against the following subject-driven generation methods: Textual Inversion \cite{textual_inversion}, Re-Imagen \cite{chenre}, BLIP-Diffusion \cite{li2024blip}, IP-Adapter \cite{Ip-adapter}, SSR-Encoder \cite{zhang2024ssr}, RPO \cite{miao2024subject}, and HybridBooth \cite{guan2024hybridbooth}.

\subsection{Implementation Details} 
Following the IP-Adapter method, we add a new image cross-attention layer to each cross-attention layer in the diffusion U-Net model. The weight of the image cross-attention $\lambda$ is set to $0.3$. $m$ is set to $1.0$. We use the IP-Adapter pre-trained model based on Stable Diffusion v1.5. {For each reference image, the IP-Adapter parameters are re-initialized from the pretrained checkpoint and optimized independently during inference, and the adapted parameters are not reused across different samples.} The DDIM scheduler step $T$ is set to $50$. We use the SGD optimizer and the learning rate is set to $0.1$. We evaluate our method using three metrics: CLIP-I, CLIP-T, and DINO-I. For CLIP-T, we calculate the CLIP text-image similarity between the generated images and the given text prompts. For CLIP-I and DINO, we first extract image features using the CLIP visual encoder and DINO, respectively. We then calculate the feature similarity between the generated images and the reference images.
All experiments are conducted on a single NVIDIA RTX3090 GPU with a random seed of $2024$.

\subsection{Qualitative Comparison}
To demonstrate the effectiveness of our method, we compare it with existing methods, including DDIM Inversion, IP-Adapter, and BLIP-Diffusion (zero-shot). For fair comparisons, we tested all models using their official codes and default hyperparameters on a single reference image. Figure \ref{fig:jungle_snow} shows the images generated with the text prompt \textit{`A S in the jungle'} and \textit{`A S in the snow'} where \textit{`S'} represents the reference image class such as \textit{`backpack, clock, ...'}. The first row shows the input reference images, while the subsequent rows display the generated images by comparison methods and our proposed method, respectively. The DDIM Inversion method generates images with the initial noise latent seed replaced by DDIM inverted latent using the same pretrained Stable Diffusion model. We can see that the IP-Adapter and BLIP-Diffusion methods struggle to generalize well to the target reference image, failing to maintain subject identity. Additionally, these methods sometimes ignore the textual prompts due to the influence of the target reference image information.
In contrast, our method performs very well in preserving subject identity, ensuring an accurate representation of the selected image subjects. Moreover, our approach achieves better text alignment compared to the baseline methods. We can see that our proposed In-Loop Model Adaptation (IMA) method can generalize well to various text prompts in Figure~\ref{fig:supp_prompts}.

\begin{table}[!ht]
    \centering
    \caption{Quantitative comparison of different methods. }
    {
    \begin{tabular}{l|ccc}
    \toprule
          Method & CLIP-T $\uparrow$ & CLIP-I $\uparrow$ & DINO $\uparrow$ \\
    \midrule
          Textual Inversion \cite{textual_inversion} & 0.255 & 0.780 & 0.569 \\
          Re-Imagen \cite{chenre} & 0.270 & 0.740 & 0.600 \\
          IP-Adapter \cite{Ip-adapter} & 0.274 & 0.809 & 0.608 \\
          BLIP-Diffusion \cite{li2024blip} & 0.282 & 0.810 & 0.660 \\          
          SSR-Encoder \cite{zhang2024ssr} & 0.308 & 0.821 & 0.612 \\
          HybridBooth \cite{guan2024hybridbooth} & 0.261 & 0.865 & 0.755\\
          RPO \cite{miao2024subject} & 0.314 & 0.833 & 0.652 \\
    \midrule
          \textbf{Ours} ($\alpha$=1.0,$\beta$=1.0) & 0.285 & 0.836 & 0.742\\
          \quad w/o IPNG ($\alpha$=1.0,$\beta$=0.0) & 0.253 & \textbf{0.894} & \textbf{0.775} \\
          \quad w/o DILG ($\alpha$=0.0,$\beta$=1.0) & \textbf{0.320} & 0.717 & 0.502 \\
    \bottomrule
    \end{tabular}
    }
    \label{tab:dreambench}
\end{table}

\subsection{Quantitative Comparisons}
In addition to qualitative comparisons, we conduct quantitative evaluations to further validate our method. We report CLIP-T, CLIP-I, and DINO scores, as shown in Table \ref{tab:dreambench}, following other methods. The experimental values for Textual Inversion \cite{textual_inversion} and Re-Imagen \cite{chenre} are referenced from BLIP-Diffusion. The results for BLIP-Diffusion and IP-Adapter are reproduced using their official codes with default hyperparameters. The results for SSR-Encoder\cite{zhang2024ssr}, RPO\cite{miao2024subject}, and HybridBooth\cite{guan2024hybridbooth} are referenced from the original paper. Using the default hyperparameters $\alpha=1.0$ and $\beta=1.0$, our method achieves an average CLIP-I score of $0.836$ and a DINO score of $0.742$ on the Dreambench benchmark, demonstrating a significant improvement in feature similarity between generated images and reference images compared to IP-Adapter. Additionally, our method shows better text alignment compared to the baseline method IP-Adapter with an average CLIP-T score of $0.285$.

Moreover, our method offers a significant advantage in computational efficiency. Unlike other finetuning-free methods that require hours of training on large domain-specific datasets, or finetuning-based methods that take tens of minutes to train on multiple target reference images with hundreds of iterations, our approach can complete training and generation in $50$ steps with a single reference image in just $13$ seconds on a single NVIDIA RTX 3090 GPU. For comparison, the vanilla text-to-image with the same Stable Diffusion model costs about $10$ seconds. This makes our method highly practical and efficient for real-world applications.

\begin{figure}
    \centering
    \includegraphics[width=\linewidth]{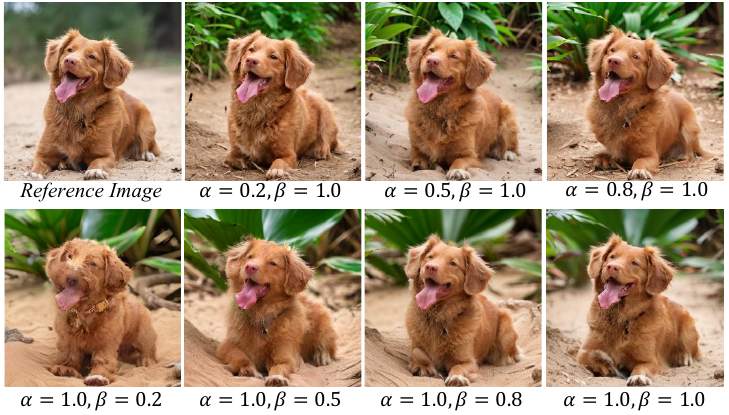}
    \caption{The generation examples with different $\alpha$ and $\beta$ values.}
    \label{fig:alpha_beta}
\end{figure}

\subsection{Ablation Study}
We also include ablation study results in Table~\ref{tab:dreambench}. As shown in the bottom rows, removing Image-Free Predicted Noise Guidance (IPNG) leads to improved alignment with the reference image—reflected by increases in CLIP-I ($0.894$) and DINO ($0.775$) scores—but results in a sharp degradation in text alignment. Conversely, removing DDIM Inversion Latent Guidance (DILG) improves the CLIP-T score ($0.320$), indicating better text alignment, while substantially degrading both CLIP-I and DINO scores. We also present a generation example in Figure~\ref{fig:alpha_beta}, illustrating the effect of varying the hyperparameters $\alpha$ and $\beta$ values in equation (\ref{eq:loss}).  As $\alpha$ increases, the alignment with the text prompt improves, while increasing $\beta$ leads to better alignment with the reference image. These ablation results reveal that both the DDIM Inversion Latent Guidance and the Image-Free Predicted Noise Guidance play an important role in the In-Loop Model Adaptation for subject-driven generation. In our experiments, we find that using the default hyperparameters $\alpha=1.0$ and $\beta=1.0$ yields competitive results across most generation scenarios. In practice, these hyperparameters can be tuned to better suit specific generation tasks or application requirements.

\subsection{Further Discussion}
\begin{figure}[!ht]
    \centering
    \includegraphics[width=\linewidth]{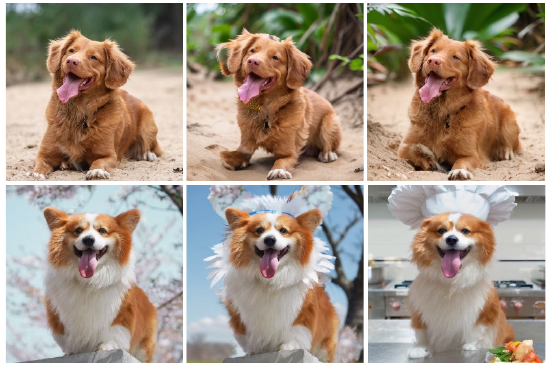}
    \caption{The generation examples with text prompts: ``\textit{a dog in the jungle}" (\textbf{Top}) and ``\textit{a dog in a chef outfit}" (\textbf{Bottom}). Reference image (\textbf{Left}), replacing the masked latent (\textbf{Middle}), and our method (\textbf{Right}).}
    \label{fig:replace_latent}
\end{figure}

In section \ref{sec:ddim}, we introduced a masked latent consistency mechanism between the DDIM inversion chain path and the generating path, to ensure that the identity of the subject from the reference image is preserved better. A natural question arises: would directly replacing the latent $X_t$ with the inverted latent $X_t^*$ in the masked region yield better results? To investigate this, we conduct a controlled experiment, illustrated in Figure \ref{fig:replace_latent}, where we apply direct latent replacement and consistency-based adaptation, respectively. The results show that this approach leads to noticeable artifacts—particularly overlapping and boundary inconsistencies around the subject—which result in less smooth visual transitions. In the second example, it even fails to align with the text prompt, highlighting the advantage of our proposed consistency-based adaptation strategy over naive latent substitution.

\begin{figure}[!ht]
    \centering
    \includegraphics[width=\linewidth]{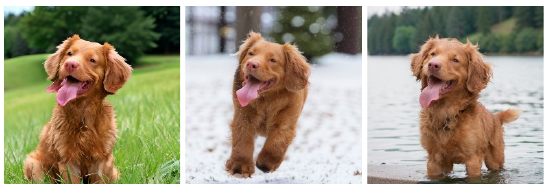}
    \caption{The generation examples for different poses with the same reference image. The text prompts are ``\textit{a dog sitting down on the grass}", ``\textit{a dog running in the snow}", and ``\textit{a dog standing in the water}", respectively.}
    \label{fig:dog_pose}
\end{figure}

We also test our method with different text prompts to assess its ability to generate images of the subject in various poses. As shown in Figure \ref{fig:dog_pose}, given the same reference image as the first image in Figure \ref{fig:replace_latent}, our approach is able to generate images with diverse poses while preserving the subject’s identity. However, there are still some limitations. For instance, it's challenging to generate images that vary significantly in viewpoint, shape, or color while maintaining identity consistency. This is mainly due to the latent consistency constraint, which limits the range of variation. We will explore other better guidance strategies, evaluation models, or agents to address this limitation in the future.

{Moreover, although a U-Net backbone is adopted in the reported experiments, the proposed method is not intrinsically tied to the U-Net architecture itself, as it performs test-time adaptation on the K/V projection parameters of the IP-Adapter rather than modifying the diffusion backbone. Consequently, in principle, the proposed method is architecture-agnostic. We consider a systematic evaluation on additional diffusion architectures, as well as exploring more adapter-agnostic design choices, to be interesting directions for future work.}

\section{Conclusion}
\label{sec:conclusion}
Subject-driven text-to-image generation is a challenging problem in computer vision. In this work, we propose an innovative inference-time In-Loop Model Adaptation method that fine-tunes the Key-Value weight parameters in the image cross-attention modules of the diffusion U-Net model, guided by coupled latent-noise consistency during the image generation process. Specifically, we introduce masked latent consistency between the DDIM inversion chain path and the generating path to enhance subject identity preservation. Additionally, we improve text prompt alignment by incorporating predicted noise regularization between the generating path and the text-to-image chain path without reference images. This coupled latent-noise loss is used to guide the in-loop model adaptation for high-fidelity subject-driven text-to-image generation. Our experimental results demonstrate the effectiveness of our proposed IMA method. In future work, we will explore better guidance for inference-time adaptation in text-to-image generation.

\section{Acknowledgment}
This work was supported by the National Natural Science Foundation of China (No. 62331014 {and No. 12426312}) and the Center for Computational Science and Engineering at Southern University of Science and Technology.

\bibliographystyle{IEEEtran}
\bibliography{IEEEabrv}

\begin{IEEEbiography}[{\includegraphics[width=1in,height=1.25in,clip,keepaspectratio]{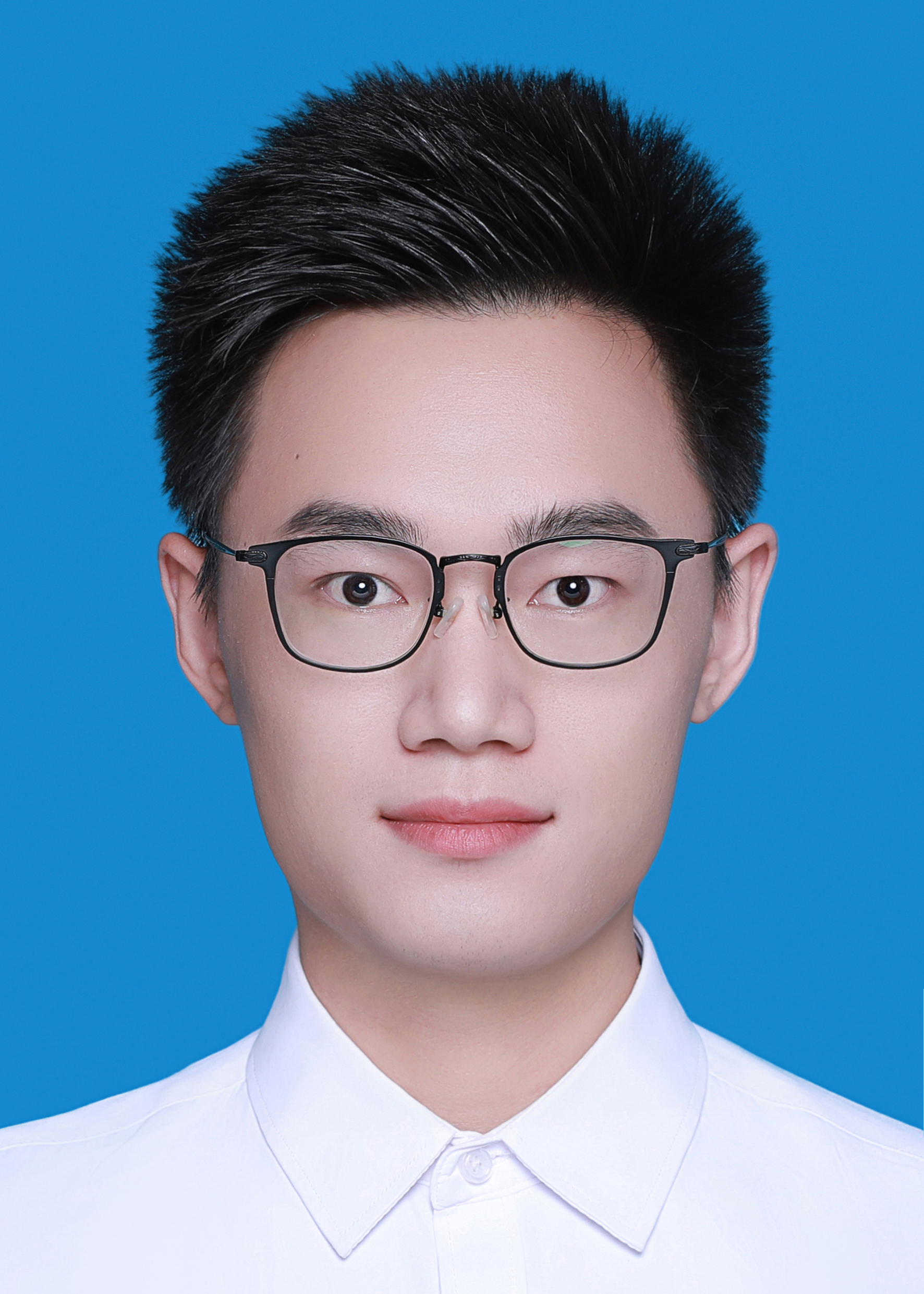}}]{Yushun Tang}
is currently a senior engineer and researcher at the TTE Lab of Huawei. He received his Ph.D. degree in Intelligent Manufacturing and Robotics from the Southern University of Science and Technology (SUSTech) in 2025. He holds a Bachelor's degree in Optoelectronic Information Science and Engineering from Harbin Engineering University (HEU) in 2019. His research focuses on Computer Vision, Transfer Learning, Domain Adaptation, Text-to-Image Generation, Reinforcement Learning, and LLM-based Agents.
\end{IEEEbiography}

\begin{IEEEbiography}
[{\includegraphics[width=1in,height=1.25in,clip,keepaspectratio]{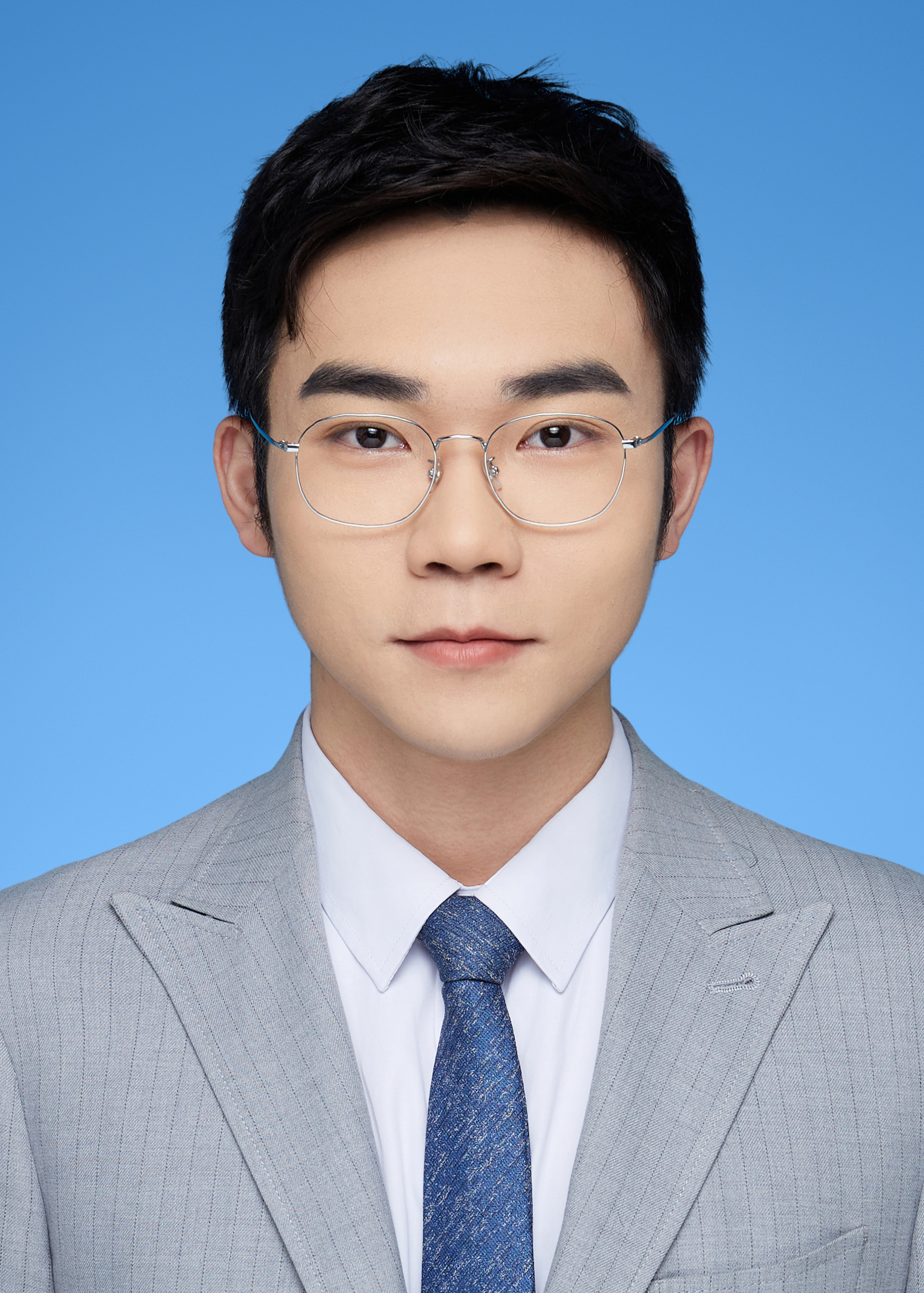}}]{Weiming Chen}
received the B.Eng. degree in mechanical design manufacture and automation, and the M.Sc. degree in electronic science and technology from Xidian University, Xi'an, China, in 2019, and 2023. He is currently pursuing the Ph.D. degree in Intelligent Manufacturing and Robotics from Southern University of Science and Technology, Shenzhen, China. His research interests include machine learning, computer vision, object detection, multi-modality, controllable text-to-image synthesis.
\end{IEEEbiography}

\begin{IEEEbiography}
[{\includegraphics[width=1in,height=1.25in,clip,keepaspectratio]{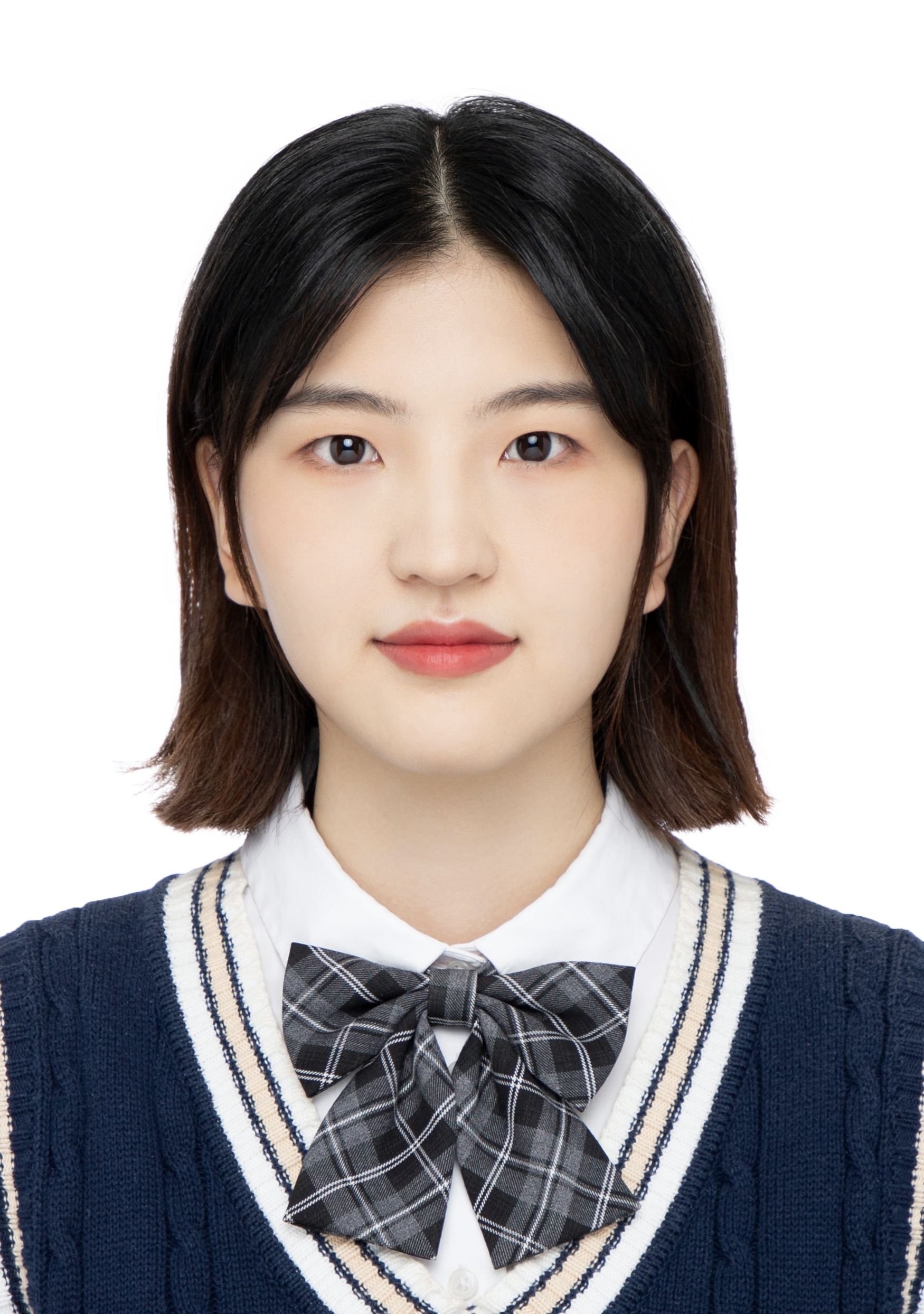}}]{Siyi Liu}
received the B. Eng. degree in Communication Engineering from Shenzhen University, Shenzhen, China, in 2023. She is currently pursuing the M. Sc. degree in Electronic Science and Technology at Southern University of Science and Technology, Shenzhen, China. Her research interests include computer vision, multi-modality, diffusion models, text-to-image generation, and image editing.
\end{IEEEbiography}

\begin{IEEEbiography}
[{\includegraphics[width=1in,height=1.25in,clip,keepaspectratio]{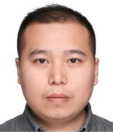}}]{Yi Zhang}
is currently an Assistant Professor in the College of Computer Science and Software Engineering at Shenzhen University. He earned his Ph.D. degree through a joint doctoral program between the Harbin Institute of Technology and the Southern University of Science and Technology (SUSTech). He received his Bachelor’s degree in Software Engineering from Northeastern University (China) and a Master’s degree in Information Systems from the University of Texas. His research interests focus on Multimodal Learning and Egocentric (first-person) vision.
\end{IEEEbiography}

\begin{IEEEbiography}
[{\includegraphics[width=1in,height=1.25in,clip,keepaspectratio]{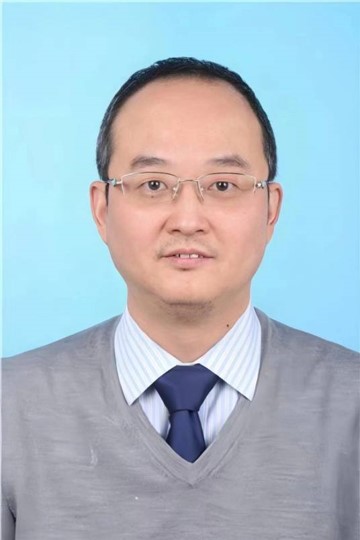}}]{Feng Wu} (Fellow, IEEE) received the BS degree in electrical engineering from Xidian University, in 1992, and the MS and PhD degrees in computer science from the Harbin Institute of Technology, in 1996 and 1999, respectively. He is currently a professor with the University of Science and Technology of China, where he is also the dean of the School of Information Science and Technology. Before that, he was a Principal Researcher and the Research Manager with Microsoft Research Asia. His research interests include image and video compression, media communication, and media analysis and synthesis. He has authored or coauthored more than 200 high quality articles (including several dozens of IEEE Transaction papers) and top conference papers on MOBICOM, SIGIR, CVPR, and ACM MM. He has 77 granted U.S. patents. His 15 techniques have been adopted into international video coding standards. As a coauthor, he received the Best Paper Award at 2009 IEEE Transactions on Circuits and Systems for Video Technology, PCM 2008, and SPIE VCIP 2007. He also received the best associate editor Award from IEEE Circuits and Systems Society, in 2012. He also serves as the TPC Chair for MMSP 2011, VCIP 2010, and PCM 2009, and the Special Sessions Chair for ICME 2010 and ISCAS 2013. He serves as an associate editor for {\sc IEEE Transactions on Circuits and Systems for Video Technology} (TCSVT), {\sc IEEE Transactions on Multimedia} (TMM), and several other international journals.
\end{IEEEbiography}

\begin{IEEEbiography}
[{\includegraphics[width=1in,height=1.25in,clip,keepaspectratio]{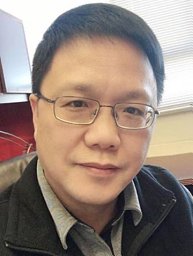}}]{Zhihai He} (Fellow, IEEE) received the B.S. degree in mathematics from Beijing Normal University, Beijing, China, in 1994, the M.S. degree in mathematics from the Institute of Computational Mathematics, Chinese Academy of Sciences, Beijing, China, in 1997, and the Ph.D. degree in electrical engineering from the University of California, at Santa Barbara, CA, USA, in 2001. In 2001, he joined Sarnoff Corporation, Princeton, NJ, USA, as a member of technical staff. In 2003, he joined the Department of Electrical and Computer Engineering, University of Missouri, Columbia, MO, USA, where he was a tenured full professor. He is currently a chair professor with the Department of Electrical and Electronic Engineering, Southern University of Science and Technology, Shenzhen, P. R. China. His current research interests include image/video processing and compression, wireless sensor network, computer vision, and cyber-physical systems.

He is a member of the Visual Signal Processing and Communication Technical Committee of the IEEE Circuits and Systems Society. He serves as a technical program committee member or a session chair of a number of international conferences. He was a recipient of the 2002 {\sc IEEE Transactions on Circuits and Systems for Video Technology} Best Paper Award and the SPIE VCIP Young Investigator Award in 2004. He was the co-chair of the 2007 International Symposium on Multimedia Over Wireless in Hawaii. He has served as an Associate Editor for the {\sc IEEE Transactions on Circuits and Systems for Video Technology} (TCSVT), the {\sc IEEE Transactions on Multimedia} (TMM), and the Journal of Visual Communication and Image Representation. He was also the Guest Editor for the IEEE TCSVT Special Issue on Video Surveillance.
\end{IEEEbiography}

\end{document}